\documentclass[10pt,twocolumn,letterpaper]{article}

\usepackage{iccv}
\usepackage{times}
\usepackage{epsfig}
\usepackage{multirow}
\usepackage{latexsym}
\usepackage{graphicx}
\usepackage{tabularx}
\usepackage{booktabs}
\usepackage{amsmath}
\usepackage{amssymb}
\usepackage{bbm}
\usepackage{xspace}
\usepackage{algorithm}
\usepackage{calrsfs}
\usepackage[noend]{algpseudocode}
\usepackage{pbox}
\usepackage{microtype} 
\usepackage[usenames,dvipsnames,svgnames,table]{xcolor}
\usepackage{subfig}
\usepackage{xspace}
\usepackage{arydshln}
\usepackage{url}

\newcommand{\shapes}{\textsc{shapes}\xspace}
\newcommand{\clevr}{\textsc{clevr}\xspace}
\newcommand{\vqa}{\textsc{vqa}\xspace}

\newcommand{\script}[1]{\footnotesize{\texttt{#1}}}
\newcommand{\myparagraph}[1]{\noindent\textbf{#1}}

\usepackage[pagebackref=true,breaklinks=true,letterpaper=true,colorlinks,bookmarks=false]{hyperref}

\iccvfinalcopy 

\def\iccvPaperID{****} 

\ificcvfinal\fi
\begin{document}

\title{Learning to Reason: End-to-End Module Networks \\ for Visual Question Answering}

\author{Ronghang Hu$^1$ \quad Jacob Andreas$^1$ \quad Marcus Rohrbach$^{1,2}$ \quad Trevor Darrell$^1$ \quad Kate Saenko$^3$ \\
$^1$University of California, Berkeley$\qquad^2$Facebook AI Research$\qquad^3$Boston University \\
{\tt\small \{ronghang,jda,trevor,rohrbach\}@eecs.berkeley.edu, saenko@bu.edu}}

\maketitle
\thispagestyle{empty}

\begin{abstract} 
Natural language questions are inherently compositional, and many are most easily answered by reasoning about 
their decomposition into modular sub-problems. For example, to answer \emph{``is there an equal number of balls 
and boxes?''} we can look for balls, look for boxes, count them, and compare the results. The recently proposed 
Neural Module Network (NMN) architecture \cite{andreas16neural,andreas2016learning} implements this approach to question answering by parsing questions into linguistic substructures and assembling question-specific deep networks from smaller modules that each solve one subtask. However, existing NMN implementations rely on brittle off-the-shelf parsers, and are restricted to the module configurations proposed by these parsers rather than learning them from data. In this paper, we propose End-to-End Module Networks (N2NMNs), which learn to reason by directly predicting instance-specific network layouts without the aid of a parser. Our model learns to generate network \emph{structures} (by imitating expert demonstrations) while simultaneously learning network \emph{parameters} (using the downstream task loss). Experimental results on the new \clevr dataset targeted at compositional question answering show that N2NMNs achieve an error reduction of nearly 50\% relative to state-of-the-art attentional approaches, while discovering interpretable network architectures specialized for each question.
\end{abstract}

\section{Introduction}

\begin{figure}[t]
\center
\includegraphics[width=\columnwidth,trim=0.0in 3.4in 1.1in 0.1in,clip]{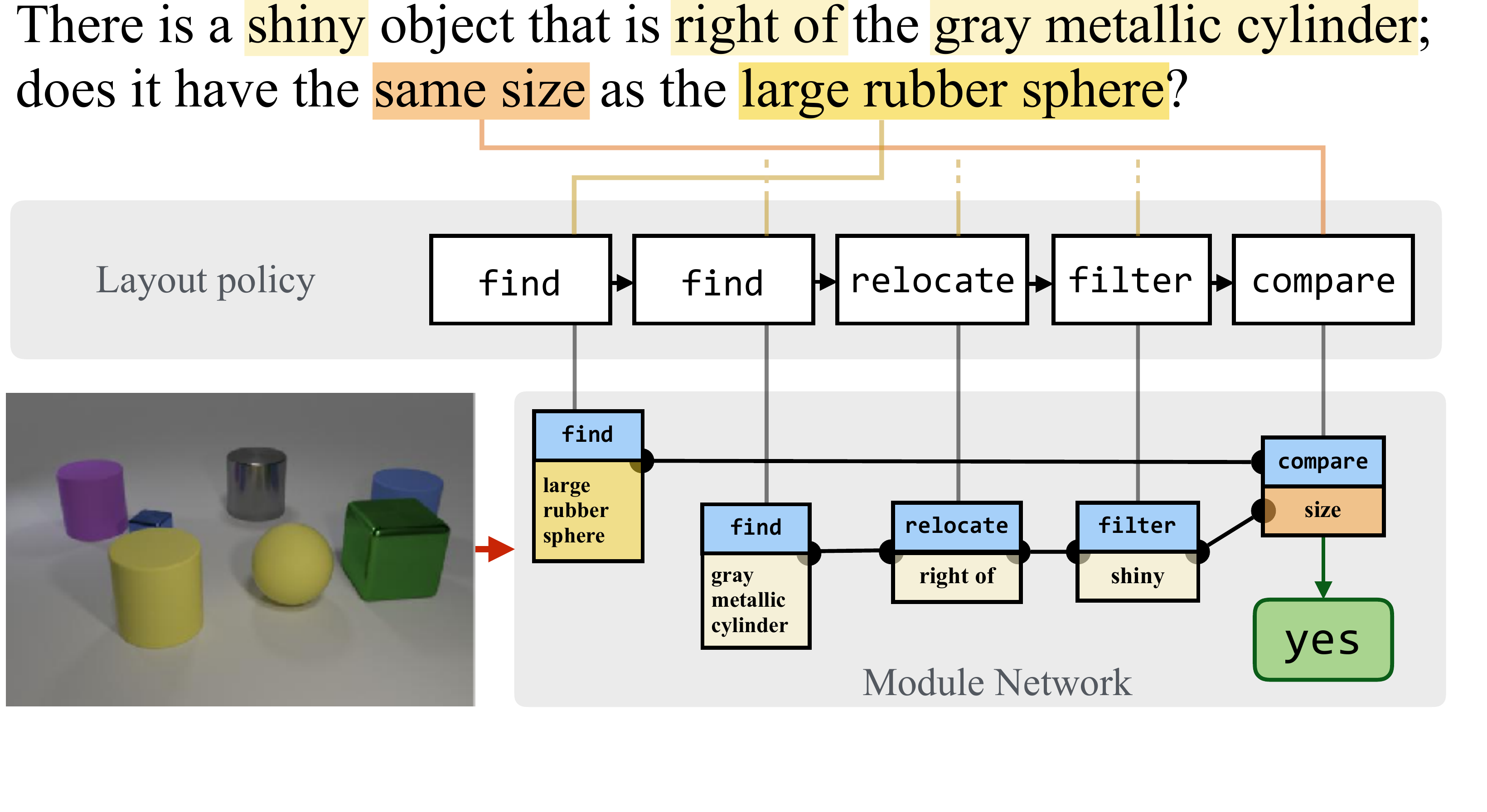}\vspace{-0.8cm}
\caption{For each instance, our model predicts a computational expression and a sequence of attentive module parameterizations. It uses these to assemble a concrete network architecture, and then executes the assembled neural module network to output an answer for visual question answering. (The
example shows a real structure predicted by our model, with text attention maps
simplified for clarity.)}
\label{fig:teaser}
\vspace{-0.5cm}
\end{figure}

Visual Question Answering (VQA) requires joint comprehension of images and text. This comprehension often depends on compositional reasoning, for example locating multiple objects in a scene and inspecting their properties or comparing them to one another (Figure \ref{fig:teaser}). While conventional deep networks have shown promising VQA performance \cite{fukui16emnlp}, there is limited evidence that they are capable of explicit compositional reasoning \cite{johnson2017clevr}. Much of the success of state-of-the-art approaches to VQA instead comes from their ability to discover statistical biases in the data distribution \cite{goyal2016vqa2}. And to the extent that such approaches are capable of more sophisticated reasoning, their monolithic structure makes these behaviors difficult to understand and explain. Additionally, they rely on the same non-modular network structure for all input questions. 

In this paper, we propose \textit{End-to-End Module Networks (N2NMNs)}: a class of models capable of predicting novel modular network architectures directly from textual input and applying them to images in order to solve question answering tasks. In contrast to previous work, our approach learns to both parse the language into linguistic structures \textit{and} compose them into appropriate layouts. 

The present work synthesizes and extends two recent modular architectures for visual problem solving. Standard neural module networks (NMNs) \cite{andreas16neural} already provide a technique for constructing dynamic network structures from collections of composable modules. However, previous work relies on an external parser to process input text and obtain the module layout. This is a serious limitation, because off-the-shelf language parsers are not designed for language and vision tasks and must therefore be modified using handcrafted rules that often fail to predict valid layouts \cite{johnson2017clevr}. Meanwhile, the compositional modular network \cite{hu2017modeling} proposed for grounding referring expressions in images does not need a parser, but is restricted to a fixed (\textit{subject, relationship, object}) structure. None of the existing methods can learn to predict a suitable structure for every input in an end-to-end manner.

Our contributions are 1) a method for learning a layout policy that dynamically predicts a network structure for each instance, without the aid of external linguistic resources at test time and 2) a new module parameterization that uses a soft attention over question words rather than hard-coded word assignments. Experiments show that our model is capable of directly predicting expert-provided network layouts with near-perfect accuracy, and even improving on expert-designed networks after a period of exploration. We obtain state-of-the-art results on the recently released \clevr dataset by a wide margin.

\section{Related work}

\begin{figure*}[t]
\centering
\vspace{-0.5cm}
\includegraphics[width=0.8\textwidth,trim=0.2in 4.2in 0.1in 0.2in,clip]{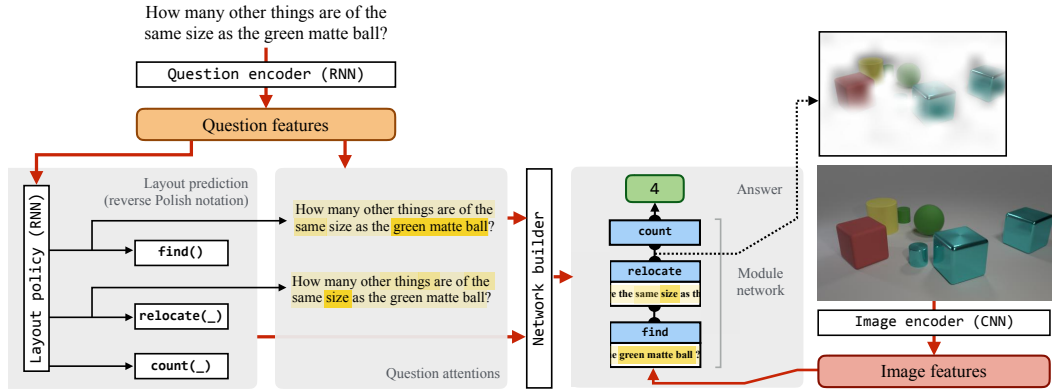}\vspace{-0.2cm}
\caption{Model overview. Our approach first computes a deep representation
   of the question, and uses this as an input to a layout-prediction policy
   implemented with a recurrent neural network. This policy emits both a 
   sequence of \emph{structural} actions, specifying a template for a modular
   neural network in reverse Polish notation, and a sequence of 	
   \emph{attentive} actions, extracting parameters for these neural modules from
   the input sentence. These two sequences are passed to a \emph{network 	    
   builder}, which dynamically instantiates an appropriate neural network
   and applies it to the input image to obtain an answer.}
\label{fig:method}
\vspace{-0.5cm}
\end{figure*}

\myparagraph{Neural module networks.} The recently proposed neural module network (NMN) architecture \cite{andreas16neural}---a general class of recursive neural networks \cite{socher2013recursive}---provides a framework for constructing deep networks with dynamic computational structure. In an NMN model, every input is associated with a \emph{layout} that provides a template for assembling an instance-specific network from a collection of shallow network fragments called \emph{modules}. These modules can be jointly trained across multiple structures to provide reusable, compositional behaviors. Existing work on NMNs has focused on natural language question answering applications, in which a linguistic analysis of the question is used to generate the layout, and the resulting network applied to some world representation (either an image or knowledge base) to produce an answer. The earliest work on NMNs \cite{andreas16neural} used fixed rule-based layouts generated from dependency parses \cite{zhu2013fast}. Later work on ``dynamic" module networks (D-NMNs) \cite{andreas2016learning} incorporated a limited form of layout prediction by learning to rerank a list of three to ten candidates, again generated by rearranging modules predicted by a dependency parse. Like D-NMNs, the present work attempts to learn an optimal layout predictor jointly with module behaviors themselves. Here, however, we tackle a considerably more challenging prediction problem: our approach learns to  optimize over the full space of network layouts rather than acting as a reranker, and requires no parser at evaluation time.

We additionally modify the representation of the assembled module networks themselves: where \cite{andreas16neural} and \cite{andreas2016learning} parameterized individual modules with a fixed embedding supplied by the parser, here we \emph{predict} these parameters jointly with network structures using a soft attention mechanism. This parameterization resembles the approach used in the ``compositional modular network'' architecture \cite{hu2017modeling} for grounding referential expressions. However, the model proposed in \cite{hu2017modeling} is restricted to a fixed layout structure of (subject, relationship, object) for every referential expression, and includes no structure search.

\myparagraph{Learning network architectures.} More generally than these dynamic / modular approaches, a long line of research focuses on generic methods for automatically discovering neural network architectures from data. Past work includes techniques for optimizing over the space of architectures using evolutionary algorithms \cite{wierstra2005modeling,floreano2008neuroevolution}, Bayesian methods \cite{bergstra2013making}, and reinforcement learning \cite{zoph2017neural}. The last of these is most closely related to our approach in this paper: both learn a controller RNN to output a network structure, train a neural network with the generated structure, and use the accuracy of the generated network to optimize the controller RNN. A key difference between \cite{zoph2017neural} and the layout policy optimization in our work is that \cite{zoph2017neural} learns a \textit{fixed} layout (network architecture) that is applied to every instance, while our model learns a layout policy that \textit{dynamically} predicts a specific layout tailored to each individual input example.

\myparagraph{Visual question answering.} The visual question answering task \cite{malinowski14nips} is generally motivated  as a test to measure the capacity of deep models to \emph{reason} about linguistic and visual inputs jointly \cite{malinowski14nips}. Recent years have seen a proliferation of datasets \cite{malinowski14nips,antol15iccv} and approaches, including models  based on differentiable memory \cite{yang2016stacked,xiong16dynamic}, dynamic prediction of question-specific computations \cite{Noh15DPPVQA,andreas2016learning}, and core improvements to the implementation of the multi-modal representation and attention mechanism \cite{fukui16emnlp,lu2016hiecoatt}. Together, these approaches have produced substantial gains over the initial baseline results published with
the first VQA datasets.

It has been less clear, however, that these improvements correspond to an improvement in the \emph{reasoning} abilities of models. Recent work has found that it is possible to do quite well on many visual QA problems by simply memorizing statistics about question / answer pairs \cite{goyal2016vqa2} (suggesting that limited visual reasoning is involved), and that models with bag-of-words text representations perform competitively against more sophisticated approaches \cite{jabri2016revisiting} (suggesting that limited linguistic 
compositionality is involved). To address this concern, newer visual question answering datasets have focused on exploring specific phenomena in compositionality and generalization; examples include the \shapes dataset \cite{andreas16neural}, the VQAv2 dataset \cite{goyal2016vqa2}, and the \clevr dataset \cite{johnson2017clevr}. The last of these appears to present the greatest challenges to standard VQA approaches and the hardest reasoning problems in general.

Most previous work on this task other than NMN uses a fixed inference structure to answer every question. However, the optimal reasoning procedure may vary greatly from question to question, so it is desirable to have inference structures that are specific to the input question. Concurrent with our work, \cite{johnson2017inferring} proposes a similar model to ours. Our model is different from \cite{johnson2017inferring} in that we use a set of specialized modules with soft attention mechanism to provide textual parameters for each module, while \cite{johnson2017inferring} uses a generic module implementation with textual parameters hard-coded in module instantiation.

\section{End-to-End Module Networks}

We propose End-to-End Module Networks (N2NMNs) to address compositionality in visual reasoning tasks. Our model consists of two main components: a set of co-attentive neural modules that provide parameterized functions for solving sub-tasks, and a layout policy to predict a question-specific layout from which a neural network is dynamically assembled. An overview of our model is shown in Figure \ref{fig:method}.

Given an input question, such as \emph{how many other things are there of the same size as the matte ball?}, our layout policy first predicts a coarse functional expression like \texttt{count(relocate(find())} that describes the structure of the desired computation, Next, some subset of function applications within this expression (here \texttt{relocate} and \texttt{find}) receive parameter vectors predicted from text (here perhaps vector representations of \emph{matte ball} and \emph{size}, respectively). Then a network is assembled with the modules according to this layout expression to output an answer.

We describe the implementation details of each neural module $f_m$ in Sec. \ref{sec:modules}, and our layout policy in Sec. \ref{sec:gen_layout}. In Sec. \ref{sec:training}, we present a reinforcement learning approach to jointly optimize the neural modules and the layout policy.

\subsection{Attentional neural modules}\label{sec:modules}

\begin{table*}[t]
\small
\centering
\vspace{-0.5cm}
\begin{tabular}{|l|c|c|c|l|}
\hline
Module name & Att-inputs & Features & Output & Implementation details \\
\hline
\texttt{find} & (none) & $x_{vis}$, $x_{txt}$ & att & $a_{out}=\mathrm{conv_2}\left(\mathrm{conv_1}(x_{vis}) \odot W x_{txt}\right)$ \\
\texttt{relocate} & $a$ & $x_{vis}$, $x_{txt}$ & att & $a_{out}=\mathrm{conv_2}\left(\mathrm{conv_1}(x_{vis}) \odot W_1\mathrm{sum}(a \odot x_{vis}) \odot W_2 x_{txt}\right)$ \\
\texttt{and} & $a_1, a_2$ & (none) & att & $a_{out} = \mathrm{minimum}(a_1, a_2)$ \\
\texttt{or} & $a_1, a_2$ & (none) & att & $a_{out} = \mathrm{maximum}(a_1, a_2)$ \\
\texttt{filter} & $a$ & $x_{vis}$, $x_{txt}$ & att & $a_{out} = \mathtt{and}(a, \mathtt{find}[x_{vis}, x_{txt}]())$, \ie reusing \texttt{find} and \texttt{and} \\
\texttt{[exist, count]} & $a$ & (none) & ans & $y=W^T \mathrm{vec}(a)$ \\
\texttt{describe} & $a$ & $x_{vis}$, $x_{txt}$ & ans & $y=W_1^T \left(W_2\mathrm{sum}(a \odot x_{vis}) \odot W_3 x_{txt}\right)$ \\
\texttt{[eq\_count, more, less]} & $a_1, a_2$ & (none) & ans & $y=W_1^T \mathrm{vec}(a_1) + W_2^T \mathrm{vec}(a_2)$ \\
\texttt{compare} & $a_1, a_2$ & $x_{vis}$, $x_{txt}$ & ans & $y=W_1^T \left(W_2\mathrm{sum}(a_1 \odot x_{vis}) \odot W_3\mathrm{sum}(a_2 \odot x_{vis}) \odot W_4 x_{txt}\right)$ \\
\hline
\end{tabular}\vspace{-0.2cm}
\caption{The full list of neural modules in our model. Each module takes 0, 1 or 2 attention maps (and also visual and textual features) as input, and outputs either an attention map $a_{out}$ or a score vector $y$ for all possible answers. The operator $\odot$ is element-wise multiplication, and $\mathrm{sum}$ is summing the result over spatial dimensions. The $\mathrm{vec}$ operation is flattening an attention map into a vector, and adding two extra dimensions: the max and min over attention map.}
\label{tab:modules}
\vspace{-0.5cm}
\end{table*}

Our model involves a set of neural modules that can be dynamically assembled into a neural network. A neural module $m$ is a parameterized function $y=f_m(a_1, a_2, \ldots ; x_{vis}, x_{txt}, \theta_m)$ that takes zero, one or multiple tensors $a_1,a_2,\ldots$ as input, using its internal parameter $\theta_m$ and features $x_{vis}$ and $x_{txt}$ from the image and question to perform some computation on the input, and outputs a tensor $y$. In our implementation, each input tensor $a_i$ is an image attention map over the convolutional image feature grid, and the output tensor $y$ is either an image attention map, or a probability distribution over possible answers.

Table \ref{tab:modules} shows the set of modules in our N2NMNs model, along with their implementation details. We assign a name to each module according to its input and output type and potential functionality, such as $\texttt{find}$ or $\texttt{describe}$. However, we note that each module is in itself merely a function with parameters, and we do not restrict its behavior during training. In addition to the input tensors (that are outputs from other modules), a module $m$ can also use two additional feature vectors $x_{vis}$ and $x_{txt}^{(m)}$, where $x_{vis}$ is the spatial feature map extracted from the image with a convolutional neural network, and $x_{txt}^{(m)}$ is a textual vector for this module $m$ that contains information extracted from the question $q$. In addition, \texttt{and} and \texttt{or} take two image attention maps as inputs, and return their intersection or union respectively.

In Table \ref{tab:modules}, the \texttt{find} module outputs an attention map over the image and can be potentially used to localize some objects or attributes. The \texttt{relocate} module transforms the input image attention map and outputs a new attention map, which can be useful for spatial or relationship inference. Also the \texttt{filter} module reuses \texttt{find} and \texttt{and}, and can be used to simplify the layout expression. We use two classes of modules to infer an answer from a single attention map: the first class has the instances \texttt{exist} and \texttt{count} (instances share the same structure, but have different parameters). They are used for simple inference by looking only at the attention map. The second class, \texttt{describe}, is for more complex inference where visual appearance is needed. Similarly, for pairwise comparison over two attention maps we also have two classes of available modules with (\texttt{compare}) or without (\texttt{eq\_count,more,less}) access to visual features.

The biggest difference in module implementation between this work and \cite{andreas16neural} is the textual component. Hard-coded textual components are used in \cite{andreas16neural}, for example, \texttt{describe}[`shape'] and \texttt{describe}[`where'] are two different instantiations that have different parameters. In contrast, our model obtains the textual input using soft attention over question words similar to \cite{hu2017modeling}. For each module $m$, we predict an attention map $\alpha_{i}^{(m)}$ over the $T$ question words (in Sec. \ref{sec:gen_layout}), and obtain the textual feature $x_{txt}$ for each module:
\begin{equation}\label{eqn:txt_input}
x_{txt}^{(m)} = \sum_{i=1}^T \alpha_i^{(m)} w_i
\end{equation}
where $w_i$ is the word embedding vector for word $i$ in the question. At runtime, the modules can be assembled into a network according to a layout $l$, which is a computation expression consisting of modules, such as $f_{m2}(f_{m4}(f_{m1}), f_{m3}(f_{m1}, f_{m1}))$, where each of $f_{m1},\cdots,f_{m4}$ is one of the modules in Table \ref{tab:modules}. 

\subsection{Layout policy with sequence-to-sequence RNN}\label{sec:gen_layout}

We would like to predict the most suitable reasoning structure tailored to each question. For an input question $q$ such as \textit{What object is next to the table?}, our layout policy outputs a probability distribution $p(l | q)$, and we can sample from $p(l | q)$ to obtain high probability layout $l$ such as \texttt{describe(relocate(find()))} that are effective for answering the question $q$. Then, a neural network is assembled according to the predicted layout $l$ to output an answer.

\begin{figure}[t]
\centering
\includegraphics[width=0.8\linewidth]{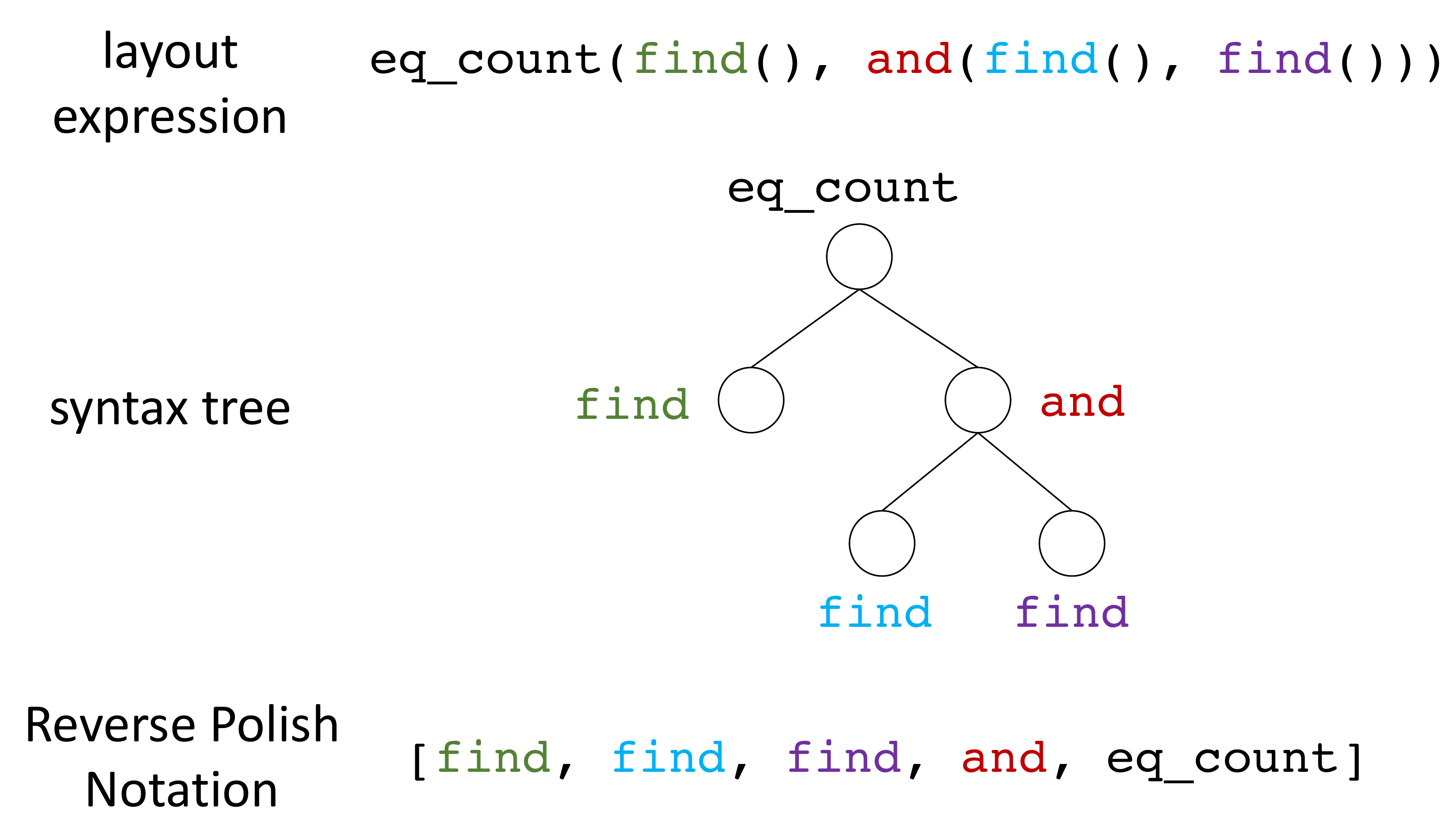}\vspace{-0.2cm}
\caption{An example showing how an arbitrary layout expression can be linearized as a sequence of module tokens.}
\label{fig:linearize}
\vspace{-0.5cm}
\end{figure}

Unlike in \cite{andreas2016learning} where the layout search space is restricted to a few parser candidates, in this work, we search over a much larger layout space: in our model, the layout policy $p(l | q ; \theta_{layout})$ predicts a distribution over the space of \textit{all possible layouts}. Every possible layout $l$ is an expression that consists of neural modules, such as $f_{m2}(f_{m1}, f_{m3}(f_{m1}, f_{m1}))$, and can be represented as a syntax tree. So each layout expression can be mapped one-to-one into a linearized sequence $l=\{m^{(t)}\}$ using Reverse Polish Notation \cite{burks54analysis} (the post-order traversal over the syntax tree). Figure \ref{fig:linearize} shows an example for an expression and its linearized module token sequence.

After linearizing each layout $l$ into a sequence of module tokens $\{m^{(t)}\}$, the layout prediction problem turns into a sequence-to-sequence learning problem from questions to module tokens. We address this problem using the attentional Recurrent Neural Network \cite{bahdanau2015neural}. First, we embed every word $i$ in the question into a vector $w_i$ (also embedding all module tokens similarly), and use a multi-layer LSTM network as the encoder of the input question. For a question $q$ with $T$ words, the encoder LSTM outputs a length-$T$ sequence $[h_1, h_2, \cdots, h_{T}]$. The decoder is a LSTM network that has the same structure as the encoder but different parameters. Similar to \cite{bahdanau2015neural}, at each time step in the decoder LSTM, a soft attention map over the input sequence is predicted. At decoder time-step $t$, the attention weights $\alpha_{ti}$ of input word at position $i \in \{1, \cdots, T\}$ are predicted as
\begin{eqnarray}
u_{ti} &=& v^T \tanh(W_1 h_i + W_2 h_t) \\
\alpha_{ti} &=& \frac{\exp(u_{ti})}{\sum_{j=1}^{T} \exp(u_{tj})} \label{eqn:att}
\end{eqnarray}
where $h_i$ and $h_t$ are LSTM outputs at encoder time-step $i$ and decoder time-step $t$, respectively, and $v$, $W_1$ and $W_2$ are model parameters to be learned from data. Then a context vector $c_t$ is obtained as $\sum_{i=1}^{T} \alpha_{ti} h_i$, and the probability for the next module token $m^{(t)}$ is predicted from $h_t$ and $c_t$ as
$
p(m^{(t)} | m^{(1)}, \cdots, m^{(t-1)}, q) = \mathrm{softmax}(W_3 h_t + W_4 c_t)
$.
We sample from $p(m^{(t)} | m^{(1)}, \cdots, m^{(t-1)}, q)$ to discretely get the next token $m^{(t)}$, and also construct its textual input $x_{txt}^{(t)}$ according to Eqn. \ref{eqn:txt_input} using the attention weights $\alpha_{ti}$ in Eqn. \ref{eqn:att}. The probability of a layout $l$ is
$
p(l | q) = \prod_{m^{(t)} \in l} p(m^{(t)} | m^{(1)}, \cdots, m^{(t-1)}, q)
$.
At test time, we deterministically predict a maximum-probability layout $l$ from $p(l | q)$ using beam search, and assemble a neural network according to $l$ to output an answer for the question.

\subsection{End-to-end training}\label{sec:training}

During training, we jointly learn the layout policy $p(l | q)$ and the parameters in each neural module, and minimize the expected loss from the layout policy. Let $\theta$ be all the parameters in our model. Suppose we obtain a layout $l$ sampled from $p(l | q; \theta)$ and receive a final question answering loss $\tilde{L}(\theta, l; q, I)$ on question $q$ and image $I$ after predicting an answer using the network assembled with $l$. Our training loss function $L(\theta)$ is as follows.
\begin{equation}\label{eqn:loss}
L(\theta) = E_{l\sim p(l | q; \theta)}[\tilde{L}(\theta, l; q, I)]
\end{equation}
where we use the softmax loss over the output answer scores as $\tilde{L}(\theta, l; q, I)$ in our implementation.

The loss function in $L(\theta)$ is not fully differentiable since the layout $l$ is discrete, so one cannot train it with full back-propagation. We optimize $L(\theta)$ using back-propagation for differentiable parts, and policy gradient method in reinforcement learning for non-differentiable part. The gradient $\nabla_\theta L$ of the loss $L(\theta)$ is
$
\nabla_\theta L = E_{l\sim p(l | q; \theta)}\left[\tilde{L}(\theta, l) \nabla_\theta \log p(l | q; \theta) + \nabla_\theta \tilde{L}(\theta, l) \right]
$
which can be estimated using Monte-Carlo sampling as
\begin{equation}\label{eqn:grad}
\nabla_\theta L \approx \frac{1}{M}\sum_{m=1}^M \left( \tilde{L}(\theta, l_m) \nabla_\theta \log p(l_m | q; \theta) + \nabla_\theta \tilde{L}(\theta, l_m) \right)
\end{equation}
where both $\log p(l_m | q; \theta)$ and $\tilde{L}(\theta, l_m)$ are fully differentiable so the above equation can be computed with back-propagation, allowing end-to-end training for the entire model. We use $M=1$ in our implementation.

To reduce the variance of the estimated gradient, we introduce a simple baseline $b$, by replacing $\tilde{L}(\theta, l_m)$ with $\tilde{L}(\theta, l_m)-b$ in Eqn. \ref{eqn:grad}, where $b$ is implemented as an exponential moving average over the recent loss $\tilde{L}(\theta, l_m)$. We also use an entropy regularization $\alpha = 0.005$ over the policy $p(l | q)$ to encourage exploration through the layout space.

\paragraph{Behavioral cloning from expert polices.}
Optimizing the loss function in Eqn. \ref{eqn:loss} from scratch is a challenging reinforcement learning problem: one needs to simultaneously learn the parameters in the sequence-to-sequence RNN to optimize the layout policy and textual attention weights to construct the textual features $x_{txt}^{(m)}$ for each module, and also the parameters in the neural modules. This is more challenging than a typical reinforcement learning scenario where one only needs to learn a policy.

On the other hand, the learning would be easier if we have some additional knowledge of module layout. While we do not want to restrict the layout search space to only a few candidates from the parser as in \cite{andreas2016learning}, we can treat these candidate layouts as an existing expert policy that can be used to provide additional supervision. More generally, if there is an expert policy $p_e(l | q)$ that predicts a reasonable layout $l$ from the question, we can first pre-train our model by behavioral cloning from $p_e$. This can be done by minimizing the KL-divergence $D_{KL}(p_e || p)$ between the expert policy $p_e$ and our layout policy $p$, and simultaneously minimizing the question answering loss $\tilde{L}(\theta, l; q, I)$ with $l$ obtained from $p_e$. This supervised behavioral cloning from the expert policy can provide a good set of initial parameters in our sequence-to-sequence RNN and each neural module. Note that the above behavioral cloning procedure is only done at training time to obtain a supervised initialization our model, and the expert policy is not used at test time.

The expert policy is not necessarily optimal, so behavioral cloning itself is not sufficient for learning the most suitable layout for each question. After learning a good initialization by cloning the expert policy, our model is further trained end-to-end with gradient $\nabla_\theta L$ computed using Eqn. \ref{eqn:grad}, where now the layout $l$ is sampled from the layout policy $p(l | q)$ in our model, and the expert policy $p_e$ can be discarded.

We train our models using the Adam Optimizer \cite{kingma2015adam} in all of our experiments. Our model is implemented using TensorFlow \cite{tensorflow2015-whitepaper} and our code is available at \url{http://ronghanghu.com/n2nmn/}.

\section{Experiments}

We first analyze our model on a relatively small \shapes dataset \cite{andreas16neural}, and then apply our model to two large-scale datasets: \clevr \cite{johnson2017clevr} and \vqa \cite{antol15iccv}.

\subsection{Analysis on the \shapes dataset}
\label{sec:exp_shapes}

The \shapes dataset for visual question answering (collected in \cite{andreas16neural}) consists of 15616 image-question pairs with 244 unique questions. Each image consists of shapes of different colors and sizes aligned on a 3 by 3 grid. Despite its relatively small size, effective reasoning is needed to successfully answer questions like \textit{``is there a red triangle above a blue shape?''}. The dataset also provides a ground-truth parsing result for each question, which is used to train the NMN model in \cite{andreas16neural}.

We analyze our method on the \shapes dataset under two settings.
In the first setting, we train our model using behavioral cloning from an expert layout policy as described in Sec. \ref{sec:training}. An expert layout policy $p_e$ is constructed by mapping the the ground-truth parsing for each question to a module layout in the same way as in \cite{andreas16neural}. Note that unlike \cite{andreas16neural}, in this setting we only need to query the expert policy at training time. At test time, we obtain the layout $l$ from the learned layout policy $p(l | q)$ in our model, while NMN \cite{andreas16neural} still needs to access the ground-truth parsing at test time.

In the second setting, we train our model without using any expert policy, and directly perform policy optimization by minimizing the loss function $L(\theta)$ in Eqn. \ref{eqn:loss} with gradient $\nabla_\theta L$ in Eqn. \ref{eqn:grad}. For both settings, we use a simple randomly initialized two-layer convolutional neural network to extract visual features from the image, trained together with other parts of our model.

\begin{table}[t]
\centering
\begin{tabular}{lc}
\toprule
Method & Accuracy \\
\hline
NMN \cite{andreas16neural} & ~~90.80\% \\
ours - behavioral cloning from expert & 100.00\% \\
ours - policy search from scratch & ~~96.19\% \\
\bottomrule
\end{tabular}\vspace{-0.2cm}
\caption{Performance of our model on the \shapes dataset. ``ours - behavioral cloning from expert'' corresponds to the supervised behavioral cloning from the expert policy $p_e$, and ``ours - policy search from scratch'' is directly optimizing the layout policy without utilizing any expert policy.}
\label{tab:results_shapes}
\vspace{-0.2cm}
\end{table}

\begin{figure}[t]
\centering
\includegraphics[width=0.95\linewidth]{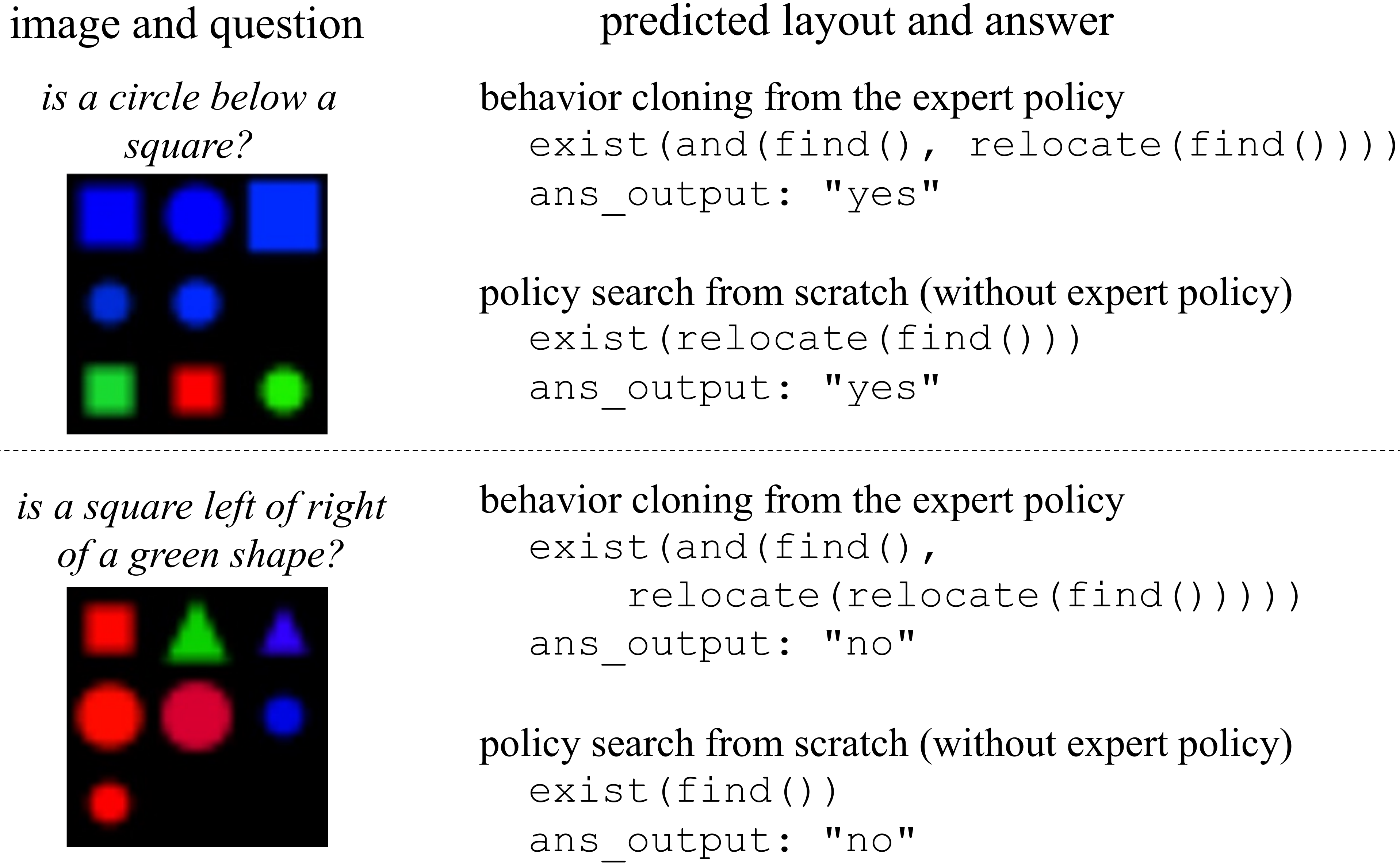}\vspace{-0.2cm}
\caption{Examples of layouts predicted by our model on the \shapes dataset, under two training settings (Sec. \ref{sec:exp_shapes}).}
\label{fig:visualization_shapes}
\vspace{-0.5cm}
\end{figure}

The results are summarized in Table \ref{tab:results_shapes}. In the first setting, we find that our model (``ours - behavioral cloning from expert'') already achieves $100\%$ accuracy. While this shows that the expert policy constructed from ground-truth parsing is quite effective on this dataset, the higher performance of our model compared to the previous NMN \cite{andreas16neural} also suggests that our implementation of modules is more effective than \cite{andreas16neural}, since the NMN is also trained with the same expert module layout obtained from the ground-truth parsing. In the second setting, our model achieves a good performance on this dataset by performing policy search from scratch without resorting to any expert policy. Figure \ref{fig:visualization_shapes} shows some examples of predicted layouts and answers on this dataset.

\begin{figure*}
\centering
\vspace{-0.5cm}
\includegraphics[height=2.5in,trim=.2in 1.5in 5.7in .2in,clip]{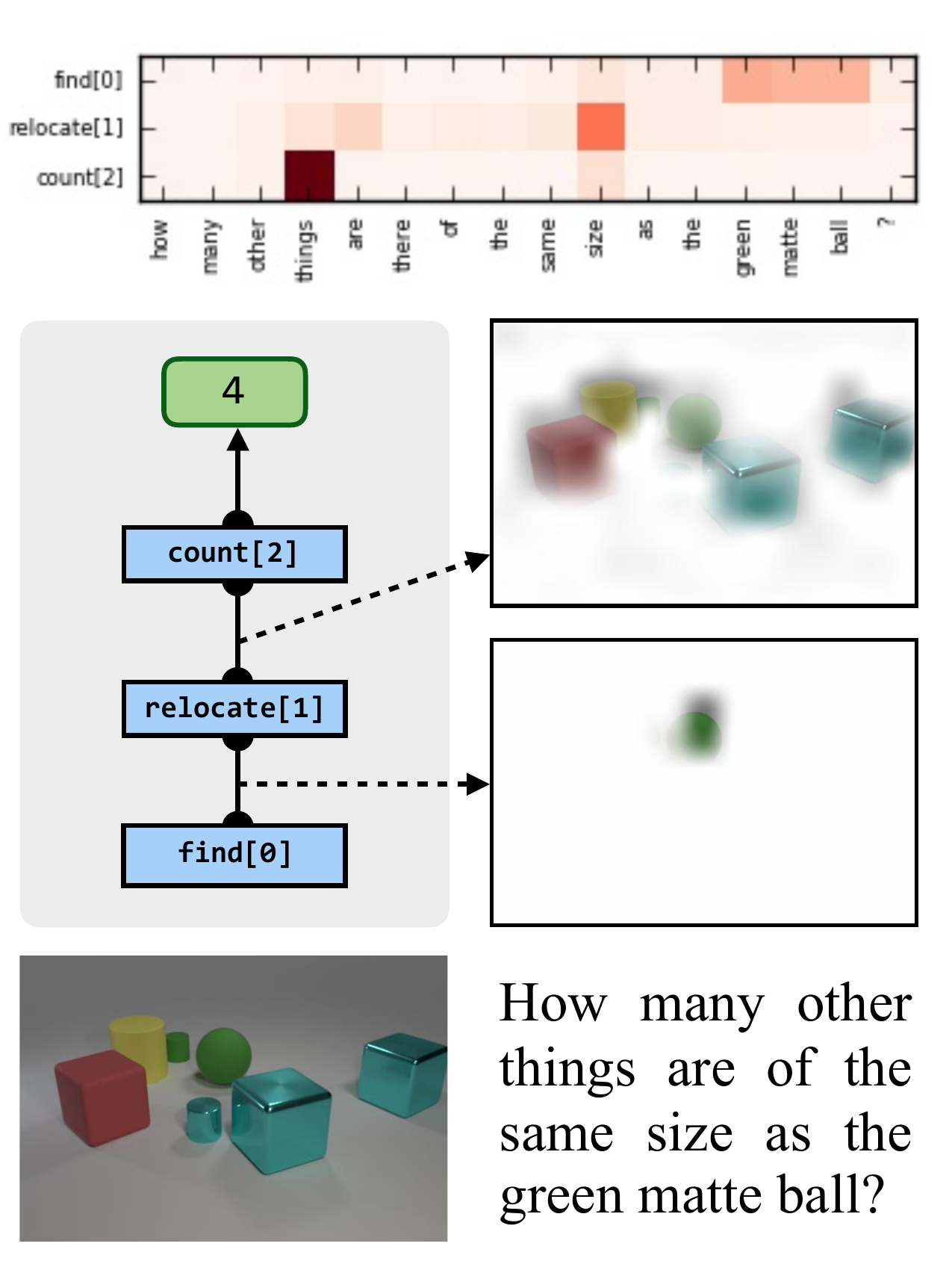}
\hspace{6em}
\includegraphics[height=2.65in,trim=0.2in 0.9in 1.8in .2in,clip]{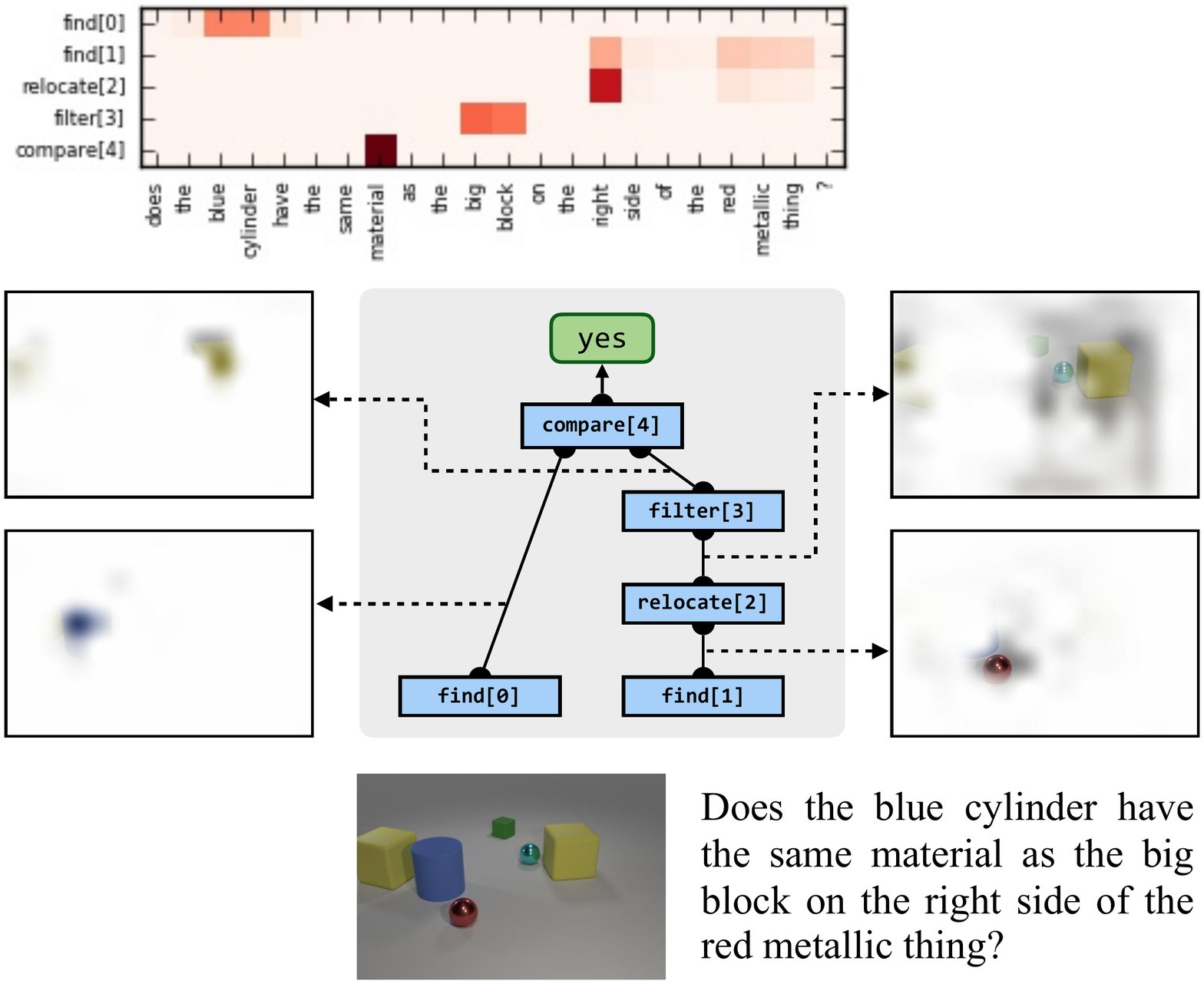}\vspace{-0.2cm}
\caption{Question answering examples on the \clevr dataset. On the left, it
can be seen that the model successfully locates the matte green ball,
attends to all the other objects of the same size, and then correctly identifies
that there are 4 such objects (excluding the initial ball). On the right, it can
be seen the various modules similarly assume intuitive semantics. 
Of particular interest is
the second \texttt{find} module, which picks up the word \emph{right} in addition
to \emph{metallic red thing}: this suggests that model can use the fact that
downstream computation will look to the right of the detected object to focus
its initial search in the left half of the image, a behavior supported by our
attentive approach but not a conventional linguistic analysis of the question.}
\label{fig:clevr_examples}
\vspace{-0.2cm}
\end{figure*}

\subsection{Evaluation on the \clevr dataset}
\label{sec:exp_clevr}

\begin{table*}[t]
\footnotesize
\center
\begin{tabular}{lcccccccccccccc}
\toprule
  & & & & \multicolumn{3}{c}{Compare Integer} & \multicolumn{4}{c}{Query Attribute} & \multicolumn{4}{c}{Compare Attribute} \\
  \cmidrule(l){5-7}\cmidrule(l){8-11}\cmidrule(l){12-15}
  Method & \textbf{Overall} & Exist & Count & equal & less & more & size & color & material & shape & size & color & material & shape \\
  \cmidrule(r){1-1} \cmidrule(r){2-2} \cmidrule(r){3-3} \cmidrule(r){4-4} \cmidrule(l){5-7}\cmidrule(l){8-11}\cmidrule(l){12-15}
  CNN+BoW \cite{zhou2016simple} & 48.4 & 59.5 & 38.9 & 50 & 54 & 49 & 56 & 32 & 58 & 47 & 52 & 52 & 51 & 52 \\
  CNN+LSTM \cite{antol15iccv} & 52.3 & 65.2 & 43.7 & 57 & 72 & 69 & 59 & 32 & 58 & 48 & 54 & 54 & 51 & 53 \\
  CNN+LSTM+MCB \cite{fukui16emnlp} & 51.4 & 63.4 & 42.1 & 57 & 71 & 68 & 59 & 32 & 57 & 48 & 51 & 52 & 50 & 51 \\
  CNN+LSTM+SA  \cite{yang2016stacked} & 68.5 & 71.1 & 52.2 & 60 & 82 & 74 & 87 & 81 & 88 & 85 & 52 & 55 & 51 & 51 \\
  \cmidrule(r){1-15}
  NMN (expert layout) \cite{andreas16neural}  & 72.1 & 79.3 & 52.5 & 61.2 & 77.9 & 75.2 & 84.2 & 68.9 & 82.6 & 80.2 & 80.7 & 74.4 & 77.6 & 79.3 \\
  \cmidrule(r){1-15}
  \multicolumn{1}{m{2.5cm}}{ours - policy search from scratch} & 69.0 & 72.7 & 55.1 & 71.6 & 85.1 & 79.0 & 88.1 & 74.0 & 86.6 & 84.1 & 50.1 & 53.9 & 48.6 & 51.1 \\
  ours - cloning expert  & 78.9 & 83.3 & 63.3 & 68.2 & 87.2 & 85.4 & 90.5 & 80.2 & 88.9 & 88.3 & 89.4 & 52.5 & 85.4 & 86.7 \\
  \multicolumn{1}{m{2.5cm}}{ours - policy search ~~~~~after cloning} & \textbf{83.7} & \textbf{85.7} & \textbf{68.5} & \textbf{73.8} & \textbf{89.7} & \textbf{87.7} & \textbf{93.1} & \textbf{84.8} & \textbf{91.5} & \textbf{90.6} & \textbf{92.6} & \textbf{82.8} & \textbf{89.6} & \textbf{90.0} \\
\bottomrule
\end{tabular}\vspace{-0.2cm}
\caption{Evaluation of our method and previous work on \clevr test set. With policy search after cloning, the accuracies are consistently improved on all questions types, with large improvement on some question types like compare color.}
\label{tab:results_clevr}
\vspace{-0.5cm}
\end{table*}

\begin{figure*}
\centering
\vspace{-0.5cm}
\small{question: \textit{do the small cylinder that is in front of the small green thing and the object right of the green cylinder have the same material?}} \\
\small{ground-truth answer: \textit{no}}
\begin{tabular}{@{}c@{}c@{}c@{}c@{}c@{}c@{}c@{}c@{}c@{}}
\footnotesize{image} & \footnotesize{layout} & \script{find[0]} & \script{relocate[1]} & \script{filter[2]} & \script{find[3]} & \script{relocate[4]} & \script{compare[5]} \\
\includegraphics[width=0.12\linewidth]{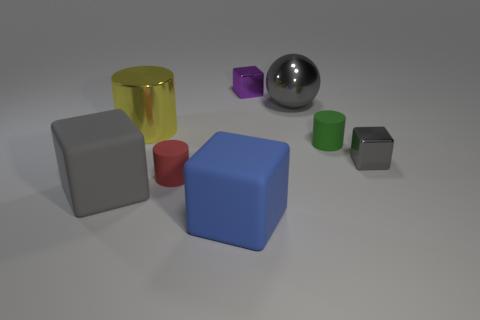} &
\includegraphics[width=0.11\linewidth]{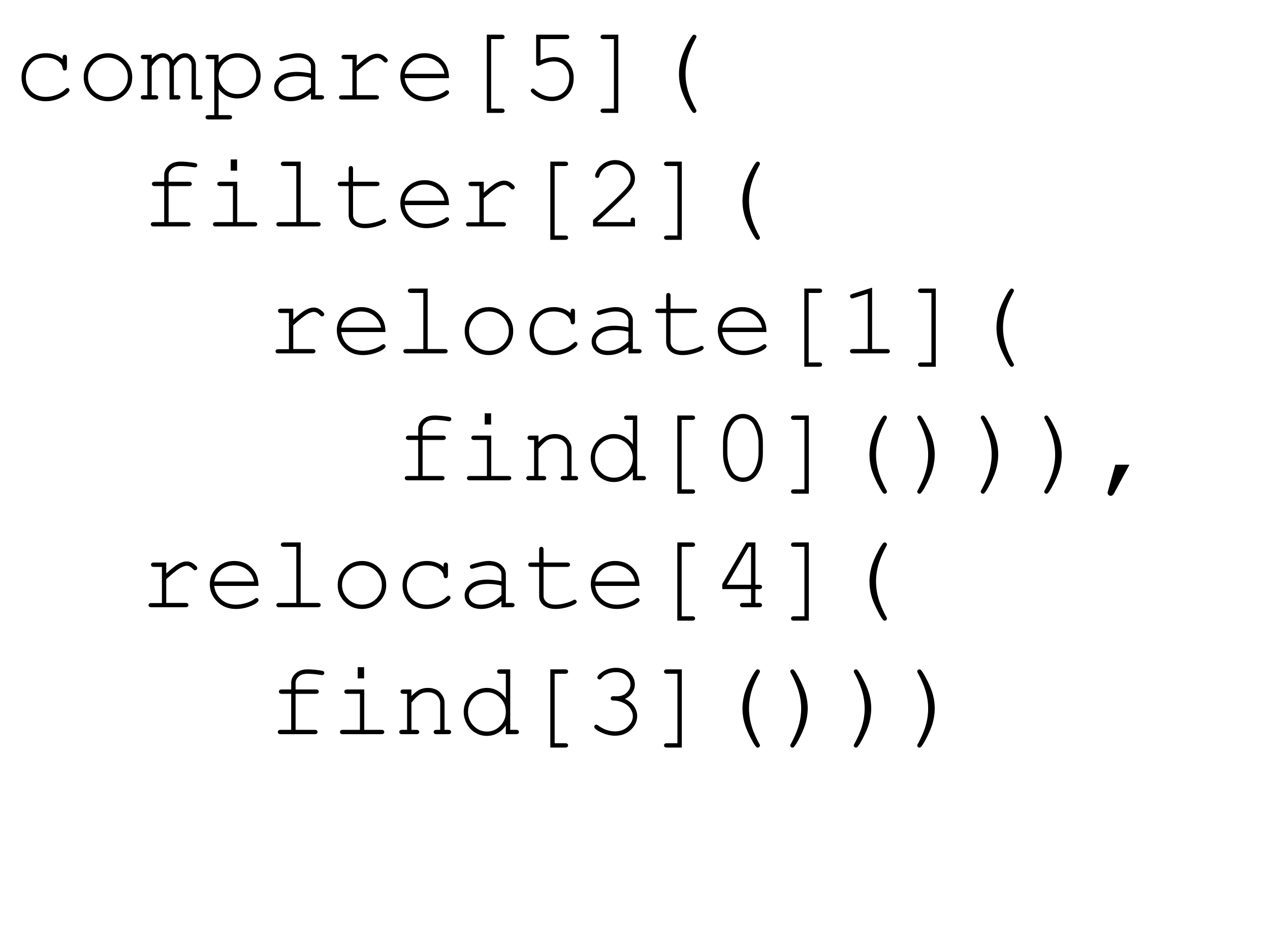} &
\includegraphics[width=0.12\linewidth]{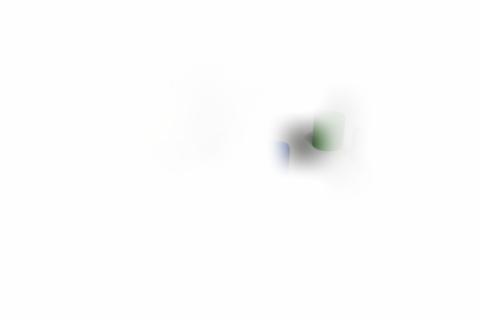} &
\includegraphics[width=0.12\linewidth]{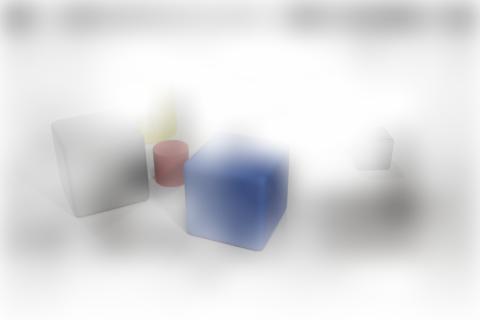} &
\includegraphics[width=0.12\linewidth]{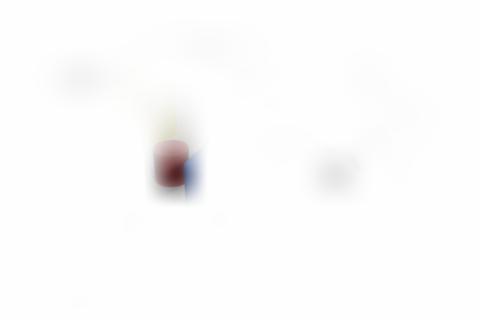} &
\includegraphics[width=0.12\linewidth]{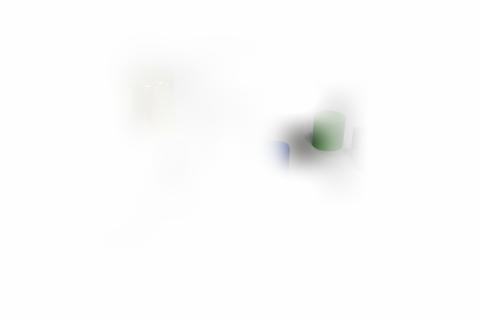} &
\includegraphics[width=0.12\linewidth]{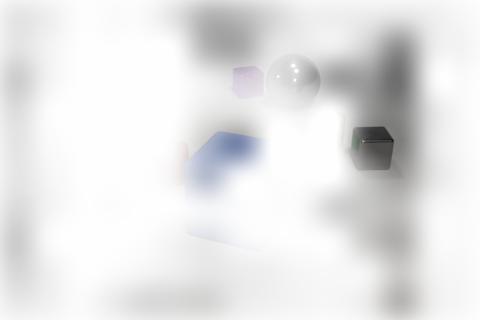} &
\includegraphics[width=0.1\linewidth]{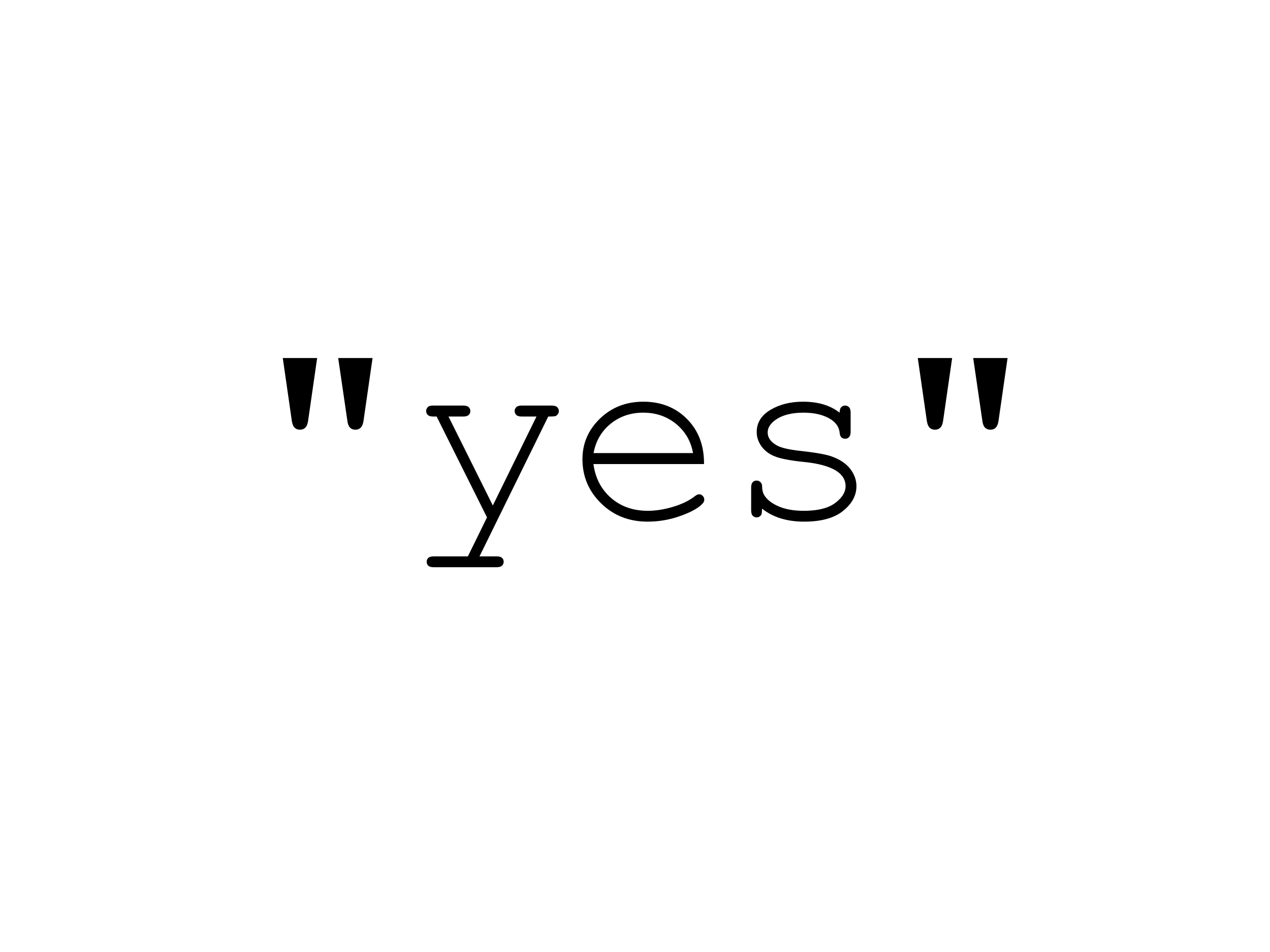} \\
\hdashline
\footnotesize{image} & \footnotesize{layout} & \script{find[0]} & \script{relocate[1]} & \script{filter[2]} & \script{find[3]} & \script{relocate[4]} & \script{filter[5]} & \script{compare[6]} \\
\includegraphics[width=0.12\linewidth]{figures/clevr/00000103_after_RL/00000103_image.jpg} &
\includegraphics[width=0.11\linewidth]{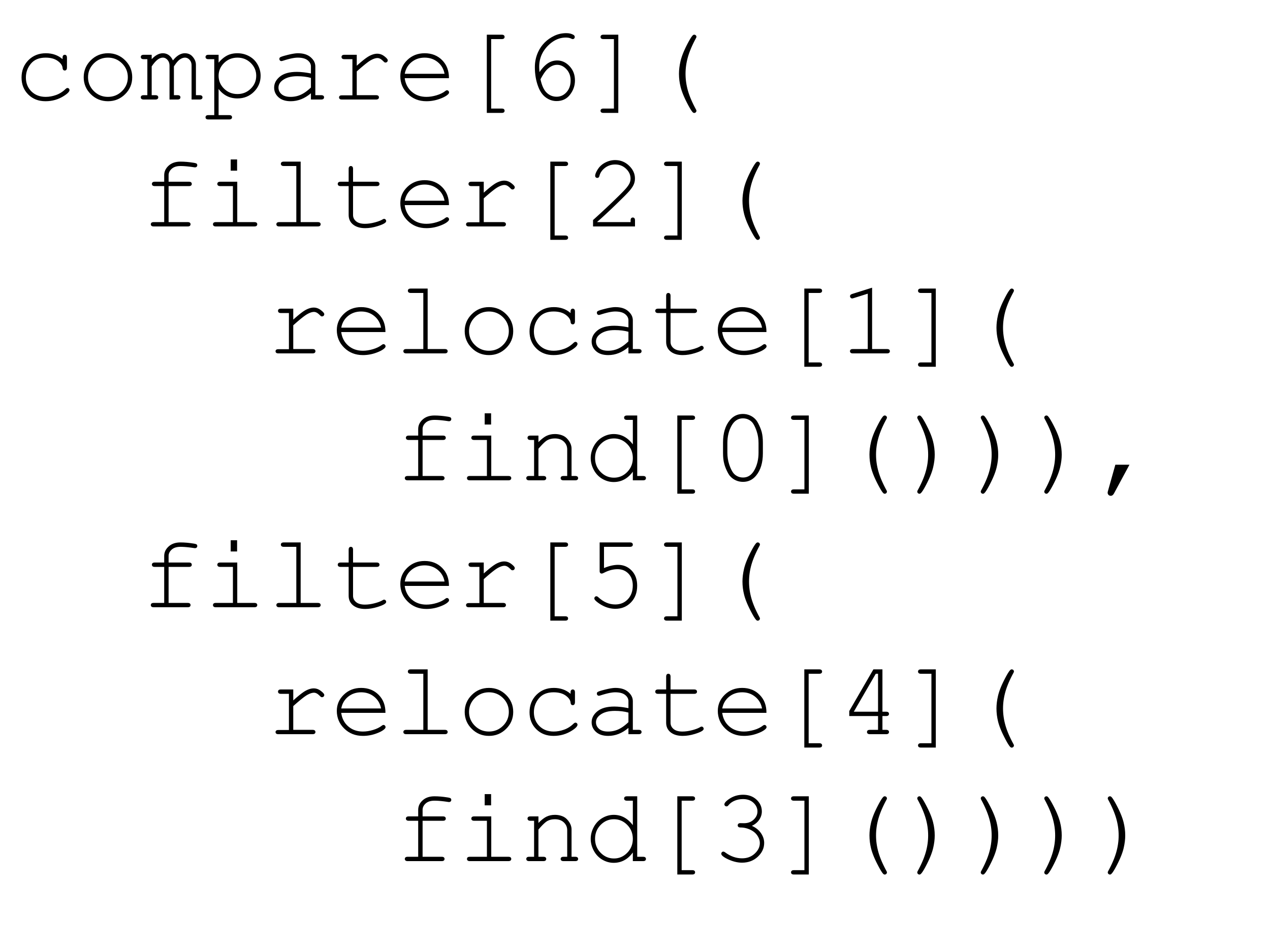} &
\includegraphics[width=0.12\linewidth]{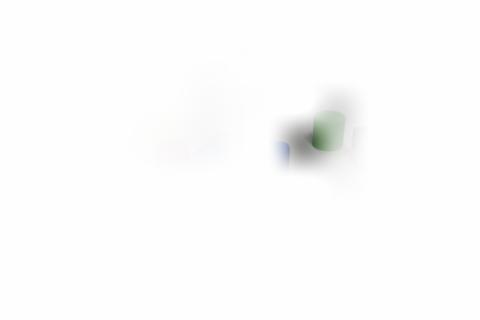} &
\includegraphics[width=0.12\linewidth]{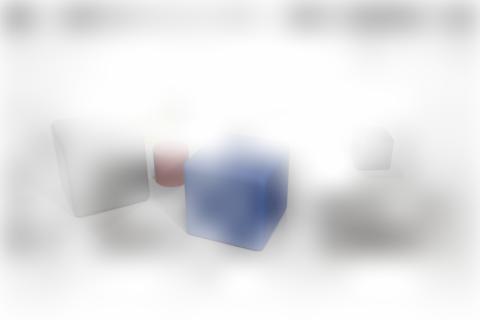} &
\includegraphics[width=0.12\linewidth]{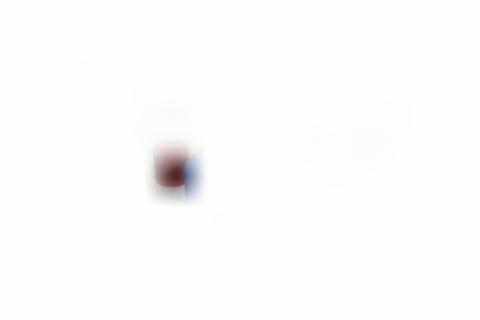} &
\includegraphics[width=0.12\linewidth]{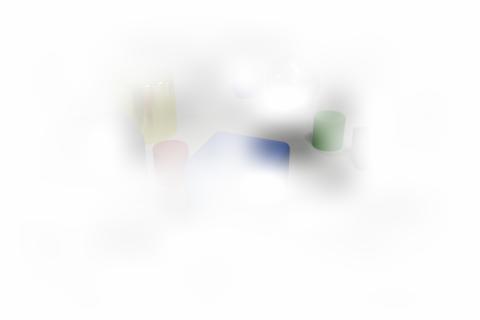} &
\includegraphics[width=0.12\linewidth]{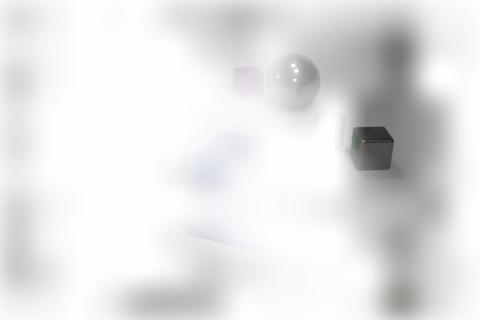} &
\includegraphics[width=0.12\linewidth]{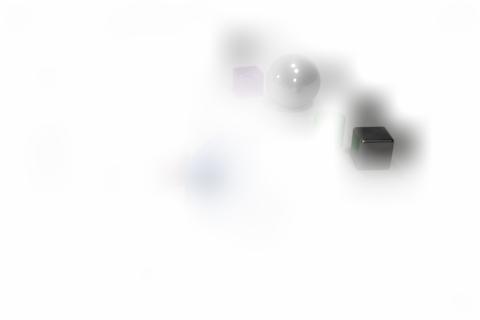} &
\includegraphics[width=0.1\linewidth]{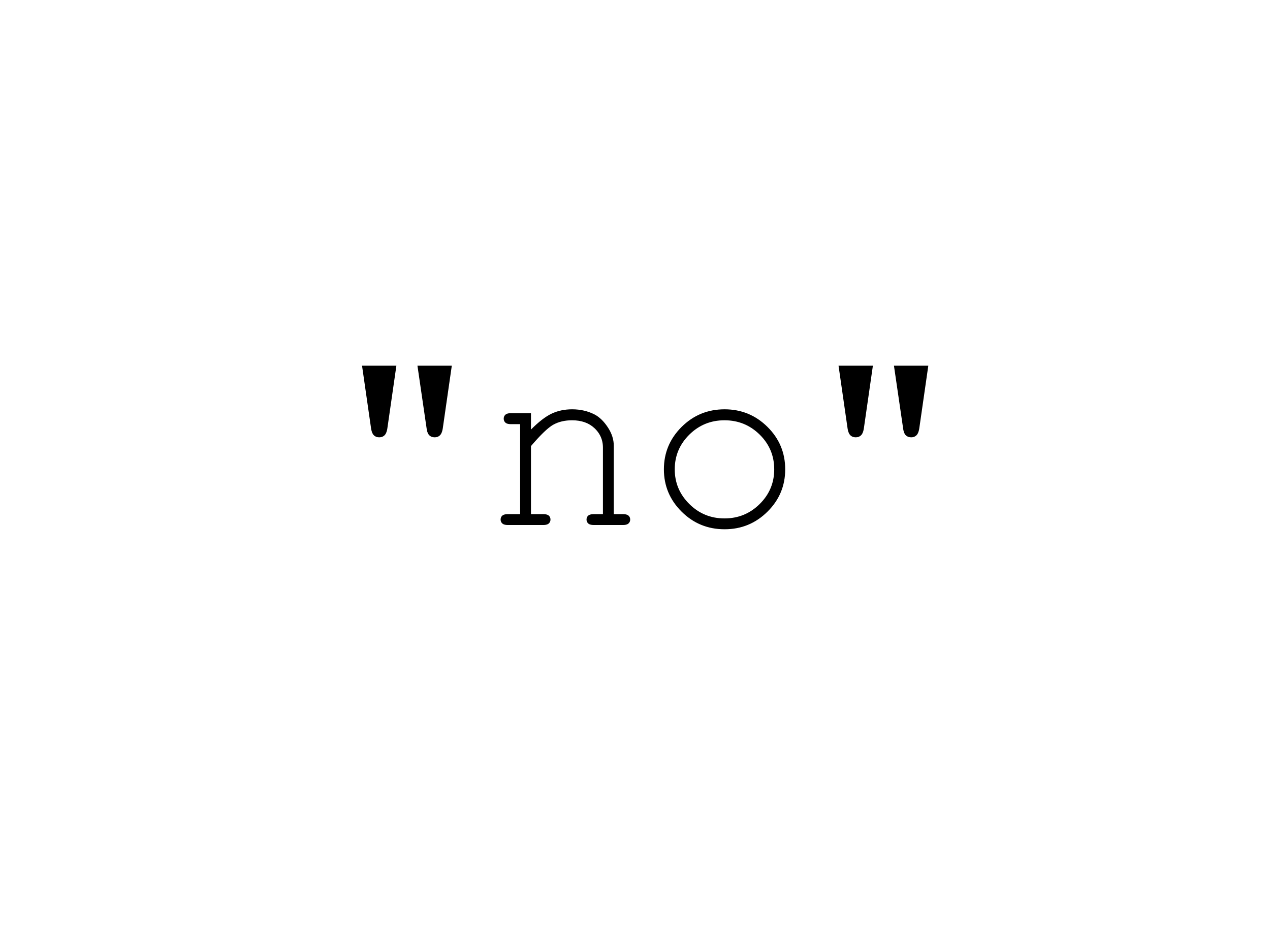} \\
\hdashline
\end{tabular}
\includegraphics[width=0.95\linewidth]{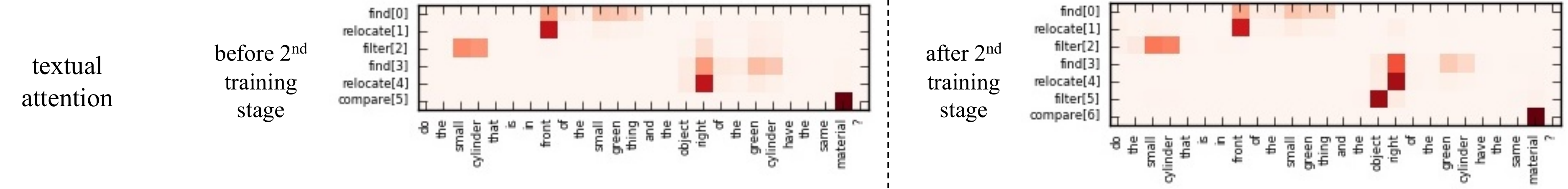}\vspace{-0.2cm}
\caption{An example illustrating the layout change before (top row) and after (middle row) the second stage of end-to-end optimization with reinforcement learning. After end-to-end learning, a new \texttt{filter} module is inserted by the layout policy to remove the attention over the non-object area before feeding it into the final \texttt{compare} module, correcting the previous error.}
\label{fig:clevr_policy_search}
\vspace{-0.2cm}
\end{figure*}

We evaluate our End-to-End Module Networks on the recently proposed \clevr dataset \cite{johnson2017clevr} with 100,000 images and 853,554 questions. The images in this dataset are photo-realistic rendered images with objects of different shapes, colors, materials and sizes and possible occlusions, and the questions in this dataset are synthesized with functional programs. Compared to other datasets for visual question answering such as \cite{antol15iccv}, the \textsc{clevr} dataset focuses mostly on the \textit{reasoning} ability. The questions in the \clevr dataset have much longer question length, and require handling long and complex inference chains to get an answer, such as \textit{``what size is the cylinder that is left of the brown metal thing that is left of the big sphere?''} and \textit{``there is an object in front of the blue thing; does it have the same shape as the tiny cyan thing that is to the right of the gray metal ball?''}.

In our experiment on this dataset, we resize each image to $480 \times 320$, and extract a $15 \times 10$ convolutional feature map from each image by forwarding the image through the VGG-16 network \cite{simonyan2015very} trained on ImageNET classification, and take the 512-channel pool5 output. To help reason about spatial properties, we add two extra $x = \frac{i}{15}$ and $y = \frac{j}{10}$ dimensions to each location $(i,j)$ on the feature map similar to \cite{hu2016segmentation}, so the final visual feature $x_{vis}$ on each image is a $15 \times 10 \times 514$ tensor. Each question word is embedded to a 300-dimensional vector initialized from scratch. We use a batch size of 64 during training.

In the first training stage, behavioral cloning is used with an expert layout policy as described in Sec. \ref{sec:training}. We construct an expert layout policy $p_e$ that deterministically maps a question $q$ into a layout $l_e$ by converting the annotated functional programs in this dataset into a module layout with manually defined rules: first, the program chain is simplified to keep all intermediate computation in  the image attention domain, and then each function type is mapped to a module in Table \ref{tab:modules} that has the same number of inputs and closest potential behavior.

While the manually specified expert policy $p_e$ obtained in this way might not be optimal, it is sufficient to provide supervision to learn good initial model parameters that can be further optimized in the later stage. During behavioral cloning, we train our model with two losses added together: the first loss is the KL-divergence $D_{KL}(p_e || p) = -\log( p(l = l_e | q) )$, which corresponds to maximizing the probability of the expert layout $l_e$ in our policy $p(l | q)$ from the sequence-to-sequence RNN, and the second loss is the question answering loss $\tilde{L}(\theta, l_e; q, I)$ for question $q$ and image $I$, where the layout $l_e$ is obtained from the expert. Note that the second loss $\tilde{L}(\theta, l_e; q, I)$ also affects the parameters in the sequence-to-sequence RNN through the textual attention in Eqn. \ref{eqn:att}.

After the first training stage, we discard the expert policy and continue to train our model for a second stage with end-to-end reinforcement learning, using the gradient in Eqn. \ref{eqn:grad}. In this stage, the model is no longer constrained to get close to the expert, but is encouraged to explore the layout space and search for the optimal layout of each question.

As a baseline, we also train our model without using any expert policy, and directly perform policy search from scratch by minimizing the loss function $L(\theta)$ in Eqn. \ref{eqn:loss}.

We evaluate our model on the test set of \clevr. Table \ref{tab:results_clevr} shows the detailed performance of our model and previous methods on each question type, where ``ours - policy search from scratch'' is the baseline using pure reinforcement learning without resorting to the expert, ``ours - cloning expert'' is the supervised behavioral cloning from the constructed expert policy in the first stage, and ``ours - policy search after cloning'' is our model further trained for the second training stage. It can be seen that without using any expert demonstrations, our method with policy optimization from scratch already achieves higher performance than most previous work, and our model trained in the first behavioral cloning stage outperforms the previous approaches by a large margin in overall accuracy. This indicates that our neural modules are capable of reasoning for complex questions in the dataset like \textit{``does the block that is to the right of the big cyan sphere have the same material as the large blue thing?''} Our model also outperforms the NMN baseline \cite{andreas16neural} trained on the same expert layout as used in our model\footnote{The question parsing in the original NMN implementation does not work on the \clevr dataset, as confirmed in \cite{johnson2017clevr}. For fair comparison with NMN, we train NMN using the same expert layout as our model.}. This shows that our soft attention module parameterization is better than the hard-coded textual parameters in NMN. Figure \ref{fig:clevr_examples} shows some question answering examples with our model.

By comparing ``ours - policy search after cloning'' with ``ours - cloning expert'' in Table \ref{tab:results_clevr}, it can be seen that the performance consistently improves after end-to-end training with policy search using reinforcement learning in the second training stage, with especially large improvement on the \textit{compare color} type of questions, indicating that the original expert policy is not optimal, and we can improve upon it with policy search over the entire layout space. Figure \ref{fig:clevr_policy_search} shows an example before and after end-to-end optimization.

\section{Evaluation on the \vqa dataset}
\label{sec:exp_vqa}

\begin{figure}
\centering
\includegraphics[width=0.7\linewidth,trim=.3in 3.5in 6.0in .4in,clip]{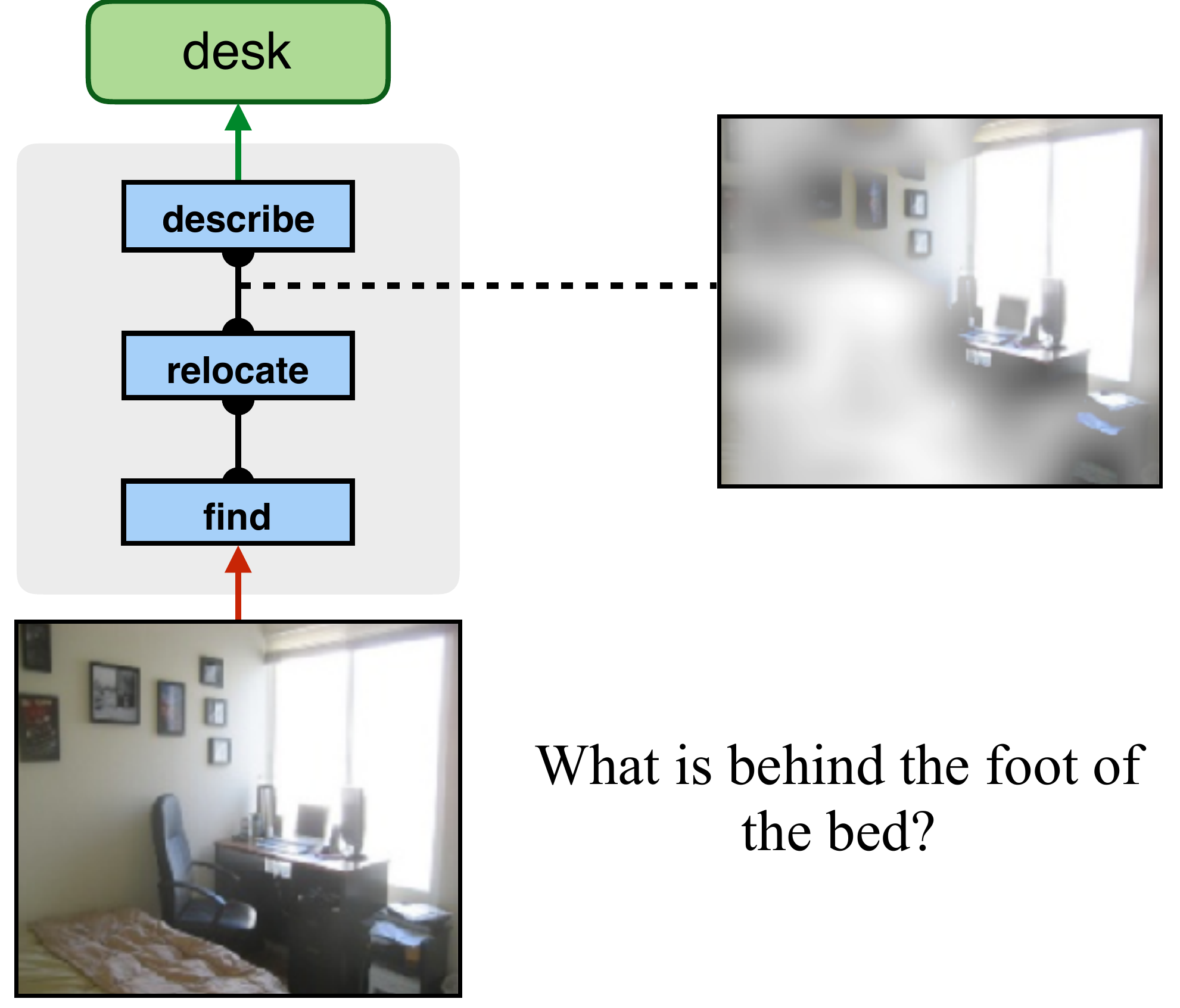}\vspace{-0.2cm}
\caption{An example from our model on the \vqa dataset.}
\label{fig:vqa}
\vspace{-0.5cm}
\end{figure}

We also evaluate our method on the \vqa dataset \cite{antol15iccv} with real images. On the \vqa dataset, although there are no underlying functional program annotation for the questions, we can still construct an expert layout policy using a syntactic parse of questions as in \cite{andreas16neural,andreas2016learning}, and train our model in the same way as in Sec. \ref{sec:exp_clevr}. We train our model using different visual features for fair comparison with other methods. Unlike previous work \cite{andreas16neural,andreas2016learning}, the syntactic parser is only used during the training stage and is not needed at test time.

\begin{table}[t]
\small
\center
\begin{tabular}{lcc}
\toprule
  Method & Visual feature & Accuracy \\
  \hline
  NMN \cite{andreas16neural} & LRCN VGG-16 & 57.3 \\
  D-NMN \cite{andreas2016learning} & LRCN VGG-16 & 57.9 \\
  MCB \cite{fukui16emnlp} & ResNet-152 & 64.7 \\
  \hline
  ours - cloning expert & LRCN VGG-16  & 61.9 \\
  ours - cloning expert & ResNet-152 & 64.2 \\
  ours - policy search after cloning & ResNet-152 & \textbf{64.9} \\
\bottomrule
\end{tabular}\vspace{-0.2cm}
\caption{Evaluation of our method on the \vqa test-dev set. Our model outperforms previous work NMN and D-NMN and achieves comparable performance as MCB.}
\label{tab:results_vqa}
\vspace{-0.5cm}
\end{table}

The results are summarized in Table \ref{tab:results_vqa} on the \vqa dataset, where our method significantly outperforms NMN \cite{andreas16neural} and D-NMN \cite{andreas2016learning} that also use modular structures, using the same LRCN VGG-16 image features (VGG-16 network fine-tuned for image captioning, as used in \cite{andreas16neural,andreas2016learning}). Compared with MCB \cite{fukui16emnlp} (the VQA 2016 challenge winner method) trained on the same ResNet-152 image features, our model achieves slightly higher performance while being more interpretable as one can explicitly see the underlying reasoning procedure. Figure \ref{fig:vqa} shows a prediction example on this dataset.

\section{Conclusion}

In this paper, we present the End-to-End Module Networks for visual question answering. Our model uses a set of neural modules to break down complex reasoning problems posed in textual questions into a few sub-tasks connected together, and learns to predict a suitable layout expression for each question using a layout policy implemented with a sequence-to-sequence RNN. During training, the model can be first trained with behavioral cloning from an expert layout policy, and further optimized end-to-end using reinforcement learning. Experimental results demonstrate that our model is capable of handling complicated reasoning problems, and the end-to-end optimization of the neural modules and layout policy can lead to significant further improvement over behavioral cloning from expert layouts.

\myparagraph{Acknowledgements.}
This work was supported by DARPA, AFRL, DoD MURI award N000141110688, NSF awards IIS-1427425, IIS-1212798 and IIS-1212928, NGA and the Berkeley Artificial Intelligence Research (BAIR) Lab. Jacob Andreas is supported by a Facebook graduate fellowship and a Huawei / Berkeley AI fellowship.

{\small
\bibliographystyle{ieee}
\bibliography{biblioLong,references,related,rohrbach}
}

\end{document}